\RequirePackage{fix-cm}
\documentclass{article}
\usepackage{iclr2027_conference,times}

\usepackage{amsmath,amssymb}
\usepackage{graphicx}
\usepackage{booktabs}
\usepackage{subcaption}
\usepackage{makecell}
\usepackage[table]{xcolor}
\usepackage[bottom]{footmisc}
\usepackage{hyperref}
\usepackage{url}
\usepackage{xspace}
\usepackage{wrapfig} %

\usepackage{amsmath,amsfonts,bm}

\def\eqref#1{equation~\ref{#1}}

\def\1{\bm{1}}

\DeclareMathAlphabet{\mathsfit}{\encodingdefault}{\sfdefault}{m}{sl}
\SetMathAlphabet{\mathsfit}{bold}{\encodingdefault}{\sfdefault}{bx}{n}

\title{SpaceCache: Intermediate Feature Reuse for Efficient Semantic Segmentation on Point Cloud Streams}

\author{Dongheon Yang\thanks{Both authors contributed equally to this work.}~, Beomseok Kang\footnotemark[1]~, Jae-Joon Kim \\
Seoul National University \\
\texttt{\{eddy49,beomseok,kimjaejoon\}@snu.ac.kr}
}

\iclrfinalcopy

\definecolor{linkblue}{HTML}{6082A3}
\hypersetup{
    colorlinks=true,
    citecolor=linkblue,
    linkcolor=linkblue,
    urlcolor=linkblue
}
\begin{document}
\maketitle
\lhead{}\renewcommand{\headrulewidth}{0pt}%

\begin{abstract}             
Point-cloud streams are central to autonomous driving and robotics, yet existing segmentation networks process every frame from scratch, although consecutive frames largely overlap. We observe that the intermediate features of point-cloud transformers remain semantically consistent across frames for the same spatial region, especially in deep stages. Building on this, we present \textbf{SpaceCache}, a training-free feature cache that reuses these features across frames. Tokens of previously observed regions skip a sequence of transformer blocks and take their cached features, only the remaining tokens are computed, and the full token layout is restored for dense prediction. Lightweight cache management keeps reused features reliable by limiting each entry’s age and evicting low-confidence entries. Unlike token merging, SpaceCache selects tokens at negligible cost and runs its cache operations once per encoder stage. On both outdoor and indoor benchmarks, such as nuScenes, Waymo, and ScanNet, with two transformer backbones of Utonia and LitePT-B, SpaceCache achieves up to 1.65× speedup with less than 1-point mIoU drop and consistently outperforms token-merging methods. The code is available at: \url{https://github.com/YangDongHeon/SpaceCache}
\end{abstract}

\section{Introduction}
\label{sec:intro}

Point-cloud segmentation is a key component in real-time systems such as autonomous vehicles and robots, where low-latency inference is critical~\citep{lang2019pointpillars, milioto2019rangenet++}. Much of the recent progress in this task has been driven by transformer-based backbones~\citep{zhao2021pt, wu2024ptv3, litept}, which treat individual points or local groups of points as tokens. However, a single LiDAR sweep often incorporates tens to hundreds of thousands of points~\citep{sun2020waymo,caesar2020nuscenes,behley2019semantickitti}, causing these backbones to process an extremely large number of tokens per frame. This cost is further compounded in point-cloud \textit{streams}, where such token-heavy inference must be repeated at every timestep. Despite this, accelerating inference in streaming point-cloud perception remains relatively underexplored.

In recent years, several studies have explored reducing inference latency in semantic segmentation, primarily for \textit{single}-frame point cloud. For example, token-merging methods~\citep{tome, algm, pitome, gitmerge3d} identify and merge similar, redundant tokens, thereby reducing the computation that would otherwise be performed on them. However, token merging requires similarity estimation and irregular aggregation operations such as \texttt{gather/reduce} to construct a new token sequence from the merged representations, introducing non-negligible GPU overhead~\citep{toma}. More recently, methods such as GitMerge3D~\citep{gitmerge3d} derive token similarities from layer-specific features, requiring merge and unmerge operations at every layer and causing this overhead to scale with network depth. Other approaches reduce computation through token pruning, which naturally avoids the repeated feature aggregation. Token Halting~\citep{ye2023efficient}, for example, progressively prunes intermediate tokens but uses an additional learned token-selection module.

Notably, temporal redundancy across \textit{multi}-frame point clouds has received relatively limited attention. One such approach, EVE~\citep{peng2021eve}, prunes redundant input points using lightweight geometric proximity across frames and reuses their previous decoder outputs, but this inevitably reduces the contextual information available to the remaining tokens. As a result, EVE also relies on an additional learned module to compensate for the resulting errors. Consequently, existing approaches either incur non-negligible overhead in token merging or rely on additional training to mitigate the contextual information loss caused by token pruning.

\begin{figure}[t]
    \centering
    \includegraphics[width=0.95\textwidth]{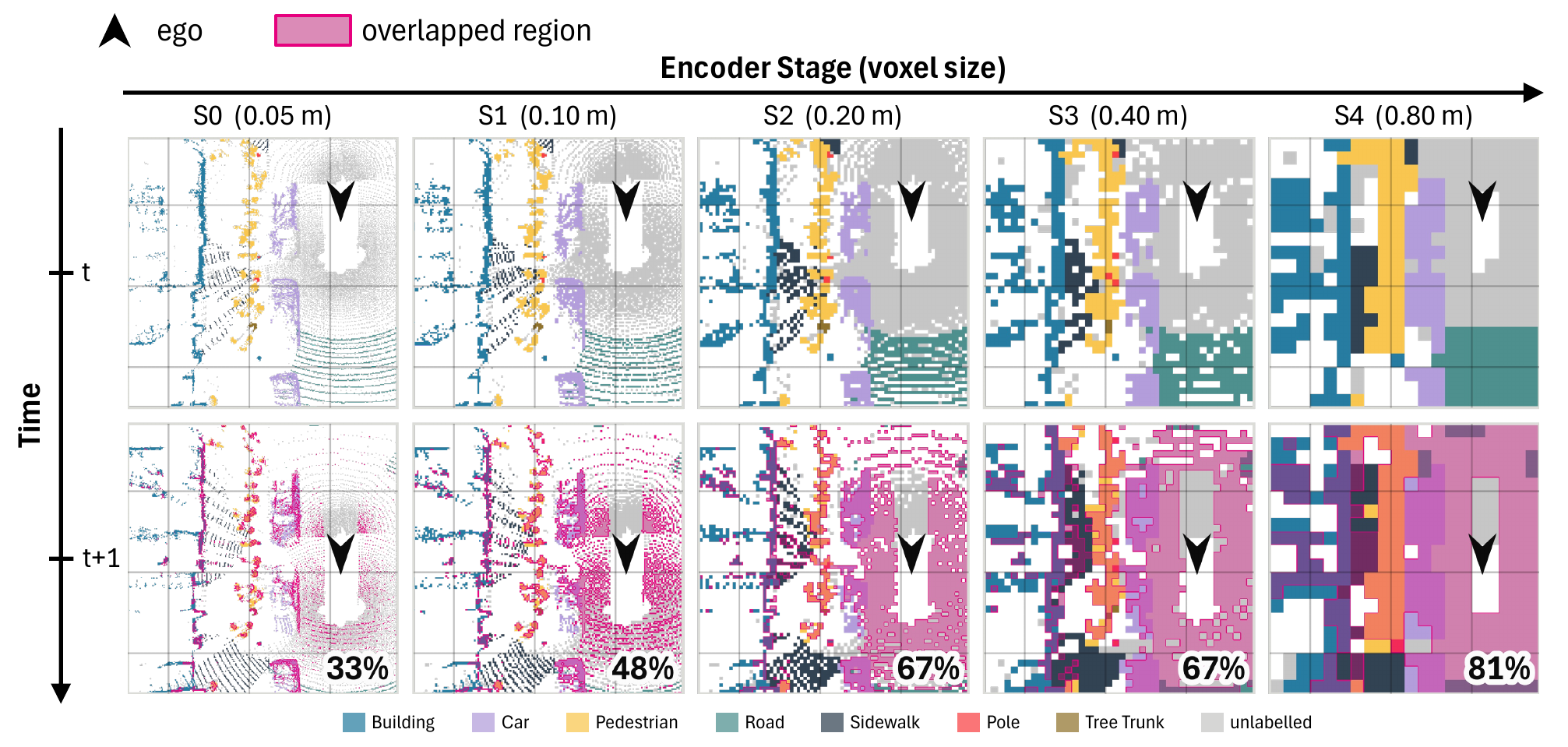}
    \vspace{-0.2cm}
    \caption{
    \textbf{Increasing temporal overlap across encoder stages.} Visualization on the Waymo dataset using the stage-wise downsampling ratios of the Utonia and LitePT-B models. Both have five encoder stages (S0--S4). %
    Each grid cell represents the voxel region corresponding to a single token, and its color indicates the ground-truth semantic class of that region. \textcolor{magenta}{Magenta} highlights voxels in frame $t{+}1$ that \textcolor{magenta}{overlap} with frame $t$. As progressive downsampling increases the voxel size at deeper stages, tokens represent larger spatial regions and exhibit higher cross-frame overlap.}
    \label{fig:temporal_overlap}
    \vspace{-0.3cm}

\end{figure}

In this paper, we introduce \textbf{SpaceCache}, an efficient intermediate-feature-reuse technique that skips computation for spatial regions observed in earlier frames and instead reuses their cached features. By reusing \textit{intermediate} token representations, namely, those at deeper encoder stages, our method achieves the low overhead of token pruning while mitigating its contextual information loss in a training-free manner. This design is motivated by two observations: at deeper encoder stages, progressive downsampling leads to higher temporal token overlap (Figure~\ref{fig:temporal_overlap}), while cached features computed in earlier frames under different contexts become increasingly semantically consistent with their corresponding current-frame features (Sec.~\ref{sec:temporal_locality}). SpaceCache targets only a few deeper encoder stages, where much of the encoder computation is concentrated, enabling substantial latency reductions. Moreover, by caching features once per stage rather than at every layer within a stage, it keeps the cache footprint to only a few MiB. Our key contributions are as follows:

\vspace{-0.2cm}
\begin{itemize}\itemsep=1pt
\item \textbf{Temporal redundancy grows with depth.} We show that deeper encoder stages exhibit higher cross-frame token overlap and feature consistency (Sec.~\ref{sec:temporal_locality}). This observation motivates SpaceCache, a training-free cache that lets tokens of previously observed regions bypass these stages.

\item \textbf{Reliability-aware cache management.} Cached features can become stale on moving objects or unreliable at ambiguous regions, degrading accuracy when they are reused. We address this by combining confidence-based commit and eviction with lifetime-based eviction (Sec.~\ref{sec:cache_management}).

\item \textbf{Better accuracy-efficiency trade-off.} Across three benchmarks in both outdoor and indoor (nuScenes, Waymo, ScanNet) and two transformer backbones (Utonia, LitePT-B), SpaceCache consistently achieves a better accuracy-efficiency trade-off than token-merging methods, reaching up to 1.65$\times$ speedup within a 1-point mIoU drop (Sec.~\ref{sec:outdoor}).
    
\end{itemize}

\begin{figure}[t]
    \centering
    \includegraphics[width=0.95\textwidth]{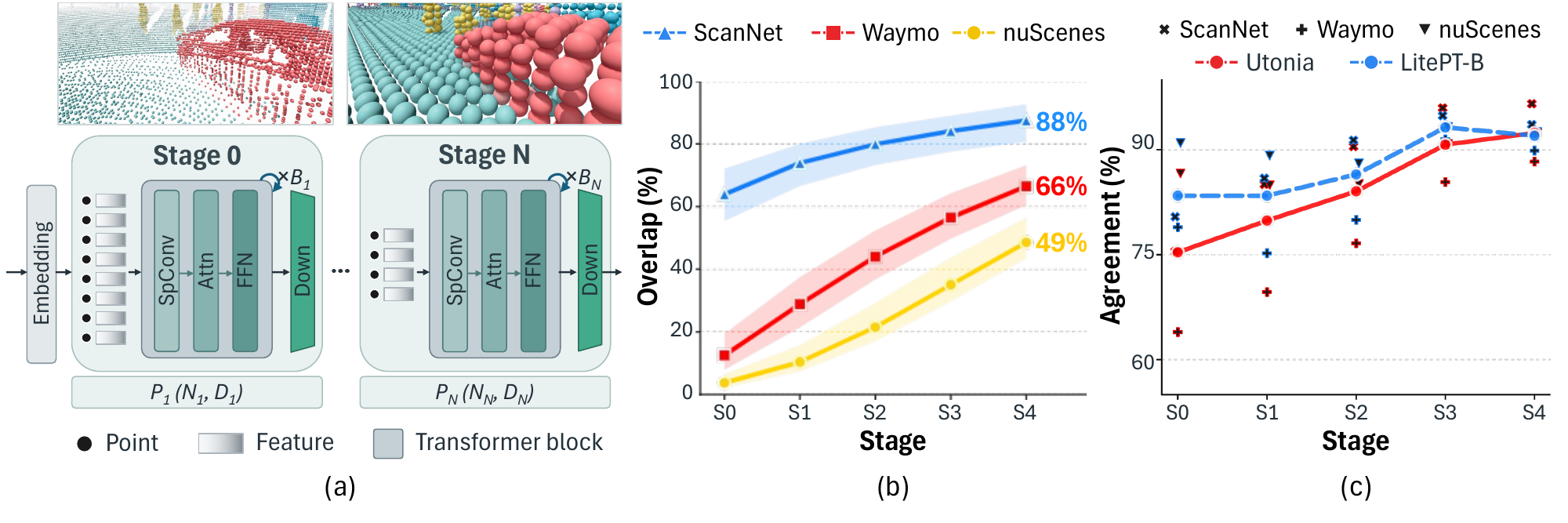}
    \vspace{-0.3cm}
    \caption{\textbf{Temporal-reuse opportunity in point-cloud encoders.}
    (a) A multi-stage encoder, where stage $l$ runs $B_l$ blocks over a fixed token set and downsampling between stages shrinks the token set while enlarging the region each token covers.
    (b) The median fraction of stage-$l$ overlapping tokens, whose region already appeared in the previous frame, grows substantially with depth.
    (c) The Reuse Agreement of stage-$l$ overlapping tokens also increases with depth, indicating that the overlapped previous-frame token and the corresponding current-frame token predict the same class (Appendix~\ref{app:reuse_agreement}).
    }
    \label{fig:motivation}
    \vspace{-0.4cm}
\end{figure}

\vspace{-0.1cm}
\section{Background and Preliminary Study}
\label{sec:motivation}

\vspace{-0.1cm}
\subsection{Network Property}

\textbf{Point-cloud encoder architectures.} Modern point-cloud encoders, such as Utonia~\citep{utonia} and LitePT~\citep{litept}, typically adopt a hierarchical multi-stage architecture, as illustrated in Figure~\ref{fig:motivation}(a). Each stage consists of a downsampling layer that aggregates multiple spatially nearby tokens into a single token, followed by $B_l$ transformer blocks that may include sparse convolution, attention, and MLP operations. An input embedding layer at the beginning of the network transforms the input point coordinates into high-dimensional feature vectors.

\textbf{Token coordinates and features.} We refer to each point or spatially aggregated representation processed by the encoder as a \emph{token}. Each token is associated with a three-dimensional coordinate and a high-dimensional feature vector, which together form the input to the encoder blocks. At stage $l$, we denote the token coordinates of block $b$ by $P_l^{(b)}{=}\{p_i^{(b)}\}_{i=1}^{N_l}$, where $N_l$ is the number of tokens and $p_i^{(b)}\in\mathbb{R}^3$ is the coordinate of token $i$. The output features are similarly denoted by $F_l^{(b)}{=}\{f_i^{(b)}\}_{i=1}^{N_l}$, where $f_i^{(b)}\in\mathbb{R}^{D_l}$ is the corresponding feature.

\textbf{Consistent token coordinates within a stage.}
The token coordinate set changes only at downsampling layers. Since the transformer blocks within a stage preserve token coordinates, they all share the same coordinate set: $P_l^{(0)}{=}P_l^{(1)}{=}\cdots{=}P_l^{(B_l)}{=}P_l$. Consequently, once tokens from different frames are matched to the same spatial region at a given stage, this correspondence remains valid throughout all blocks in that stage. This property enables intermediate features to be reused across the entire sequence of blocks, rather than only at an individual block.

\vspace{-0.1cm}
\subsection{Feature Reuse Opportunity}
\label{sec:temporal_locality}

Our strategy for accelerating streaming point-cloud perception is to reuse intermediate features from previously observed spatial regions rather than repeatedly encoding them. Effective feature reuse depends on two questions: (1) how many tokens overlap across consecutive frames, and (2) whether the features of these overlapping tokens can be reliably reused. We examine these questions through temporal overlap and semantic consistency of intermediate features across frames.

\textbf{Temporal overlap of intermediate tokens.} Feature reuse is beneficial only when many tokens correspond to spatial regions already observed in previous frames. We therefore quantify their overlap across consecutive frames at each encoder stage. Because progressive downsampling changes the spatial granularity represented by each token, cross-frame overlap can vary across stages.

To quantify how often tokens from different frames correspond to the same physical regions, we transform their coordinates into a shared global coordinate system so that they can be compared in a common spatial reference frame (details; Sec.~\ref{sec:cache_operation}). Figure~\ref{fig:motivation}(b) shows that temporal overlap increases substantially with network depth: from 4\% to 49\% on nuScenes, 13\% to 66\% on Waymo, and 64\% to 88\% on ScanNet. Progressive downsampling makes deeper tokens represent larger spatial regions, increasing the likelihood that the same regions are observed across consecutive frames.

\textbf{Semantic consistency of overlapping features.} Spatial overlap alone does not guarantee reliable feature reuse; features from overlapping regions must also preserve consistent semantic information across frames. In other words, a feature computed for a region in a previous frame should remain representative of the same semantic category when reused in the current frame.

To quantify this consistency, we define ``Reuse Agreement". For each semantic category $c$, we first construct a mean feature $\mu_c$ by averaging the features $f_i$ associated with that category. We then measure whether the reused feature and its corresponding current-frame feature are associated with the same semantic category (details; Appendix~\ref{app:reuse_agreement}). Figure~\ref{fig:motivation}(c) shows that Reuse Agreement increases with encoder depth, from approximately 75--83\% at Stage~0 to 91--93\% at Stages~3--4. In particular, the agreement begins to saturate from Stage~3, indicating that reused features at deeper stages largely preserve the semantic information of their corresponding current-frame features.

\textbf{Design Takeaway.}
Figures~\ref{fig:motivation}(b) and (c) show that deeper encoder stages exhibit both higher temporal overlap and stronger semantic consistency across frames. These properties make deeper intermediate representations increasingly suitable for temporal feature reuse.

\section{Method}
\label{sec:method}

\subsection{Overview of SpaceCache}
\label{sec:overview}

\begin{figure}[t]
    \centering
    \includegraphics[width=0.95\textwidth]{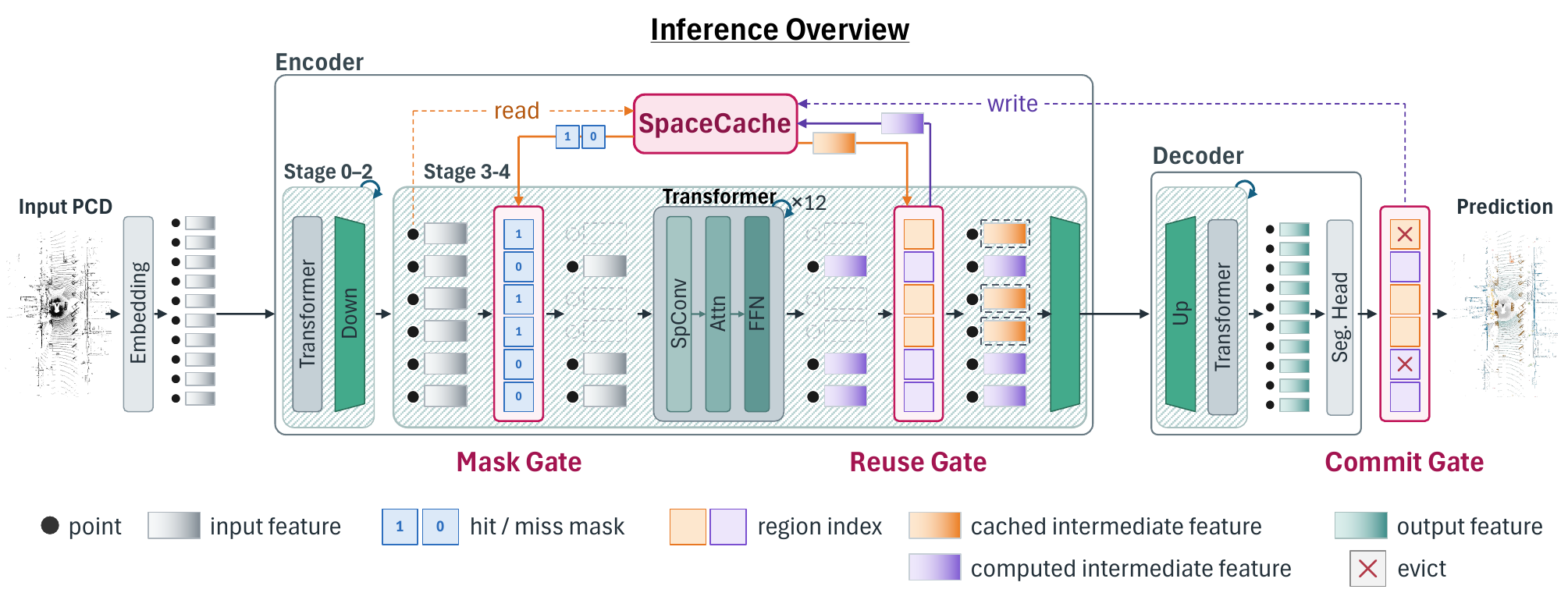}
    \vspace{-0.3cm}
    \caption{\textbf{Overview of SpaceCache.} Inference pipeline with the Mask, Reuse, and Commit Gates. Cache-hit tokens bypass the selected transformer blocks, while cache-miss tokens are computed through the blocks. The Reuse Gate restores the full token layout, and the Commit Gate updates the cache after prediction. Each cached stage has its own SpaceCache.}
    \label{fig:overview}
    \vspace{-0.3cm}
\end{figure}

Figure~\ref{fig:overview} illustrates the overall inference flow of a point-cloud encoder integrated with SpaceCache. SpaceCache stores reliable intermediate features from the \textit{last} transformer block of selected encoder stages and reuses them when the corresponding spatial regions are observed again (cache-hit) in subsequent frames. The process is implemented through three gates with distinct roles: the Mask Gate, Reuse Gate, and Commit Gate.

The \textbf{Mask Gate} queries SpaceCache using the coordinates $p_i$ of current-frame tokens to identify cache hits and misses. Cache-hit tokens (dotted bars in Figure~\ref{fig:overview}) bypass the transformer blocks, while only cache-miss tokens are processed. After the transformer blocks, the \textbf{Reuse Gate} combines the newly computed features of cache-miss tokens (purple bars) with the cached features of cache-hit tokens (orange bars) and restores the full-token layout for subsequent layers. Newly computed cache-miss features are also retained as candidates for the cache update. Once the frame prediction is produced, the \textbf{Commit Gate} determines which cached entries and candidate features remain valid and updates SpaceCache accordingly.

\vspace{-0.1cm}
\subsection{Operation Details of SpaceCache}
\label{sec:cache_operation}

\begin{figure}[t]
    \centering
    \includegraphics[width=0.95\textwidth]{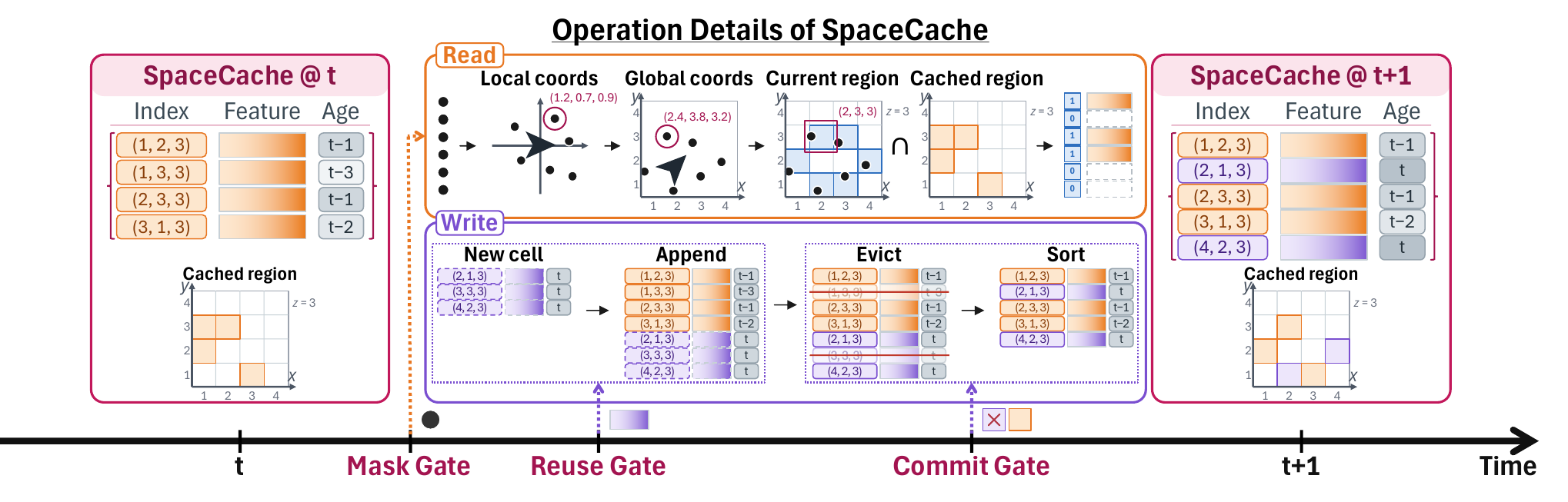}
    \vspace{-0.3cm}
    \caption{\textbf{Operation details of SpaceCache.} Each entry stores a region index, cached feature, and age. \textbf{Read:} Current-frame tokens are mapped to global region indices and queried at the Mask Gate to identify cache hits; the corresponding cached features are then retrieved for reuse at the Reuse Gate. \textbf{Write:} Newly computed cache-miss features become candidate entries at the Reuse Gate; after prediction, the Commit Gate evicts invalid entries, commits valid candidates, and re-sorts the cache for the next-frame lookup.}
    \label{fig:operation_detatil}
    \vspace{-0.3cm}    
\end{figure}

\textbf{SpaceCache components.} Figure~\ref{fig:operation_detatil} illustrates the operations of SpaceCache in detail. Each entry in the cache consists of three components: a \textit{Regional Index} $\mathbf{v}\in\mathbb{Z}^3$ recording a previously observed spatial region, a \textit{Feature} $\mathbf{f}\in\mathbb{R}^{n}$ storing the intermediate representation computed for that region, and a scalar \textit{Age} $a\in\mathbb{N}$ counting the frames since the entry was inserted. The entries are maintained in arrays sorted by the index, allowing each lookup to be efficiently performed by binary search.

\textbf{Transformation from token coordinate to region index.} To reuse features across frames, SpaceCache must determine whether a token corresponds to a previously observed spatial region. However, token coordinates are generally expressed in each frame's \textit{local} coordinate system and hence cannot be directly compared across time. We therefore transform each token coordinate $p_i^t$ into a shared \textit{global} coordinate system using the ego pose $T_t$ and then quantize it into a spatial cell whose size depends on the encoder stage, so that observations of the same physical region across frames are assigned the same region index: $\bar{p}_i^t = T_t p_i^t$ and $v_i^t = \lfloor \bar{p}_i^t / s_l \rfloor$,
where $s_l=2^l g$ is the cell size at stage $l$, $g$ is the input voxel size, and $2^l$ is the corresponding downsampling ratio. Finally, the three-dimensional region index $v_i^t$ is mapped into a scalar key $\phi(v_i^t)$ for efficient cache lookup on the GPU (Appendix~\ref{app:key_mapping}). 

\textbf{Step 1: cache read.} At the Mask Gate, SpaceCache is queried using the region indices of the current-frame tokens. The lookup returns a cache-hit mask indicating which regions are already present in the cache (denoted as ``Current region $\cap$ Cache region" in Figure~\ref{fig:operation_detatil}), together with the cached features of the cache-hit tokens. The Mask Gate uses the hit mask to bypass redundant computation, while the Reuse Gate later retrieves the corresponding cached features (orange bars in ``Read" of the figure). This operation is lightweight because it only requires checking whether each region index exists in the cache. Since the cache entries are sorted by region indices, we perform a binary search for each token in parallel on the GPU. Each lookup takes $\mathcal{O}(\log M)$ operations, resulting in $\mathcal{O}(N\log M)$ total work for $N$ input tokens and $M$ cached entries.

\textbf{Step 2: candidate write.} At the Reuse Gate, newly computed features of cache-miss tokens are paired with their region indices and temporarily stored as \textit{candidate} entries (purple bars in ``Write" of the figure). These candidates remain separate from the active cache until the frame-level semantic predictions become available for reliability assessment. Meanwhile, the newly computed features are combined with the cached features of cache-hit tokens to restore the original token layout. The candidates are then passed to the Commit Gate for validation and possible insertion into the cache.

\vspace{-0.1cm}
\textbf{Step 3: cache commit and eviction.} Intermediate feature reuse does not guarantee that a cached feature remains valid for every subsequent frame. First, cached features may become stale as the scene changes, particularly in regions containing dynamic objects, where the semantic content of the same spatial region can change across frames. Second, a feature may be unreliable when it is first computed, particularly in ambiguous regions such as object boundaries or when reduced context leads to uncertain predictions. Repeatedly reusing such features can propagate errors across subsequent frames. Thus, SpaceCache evaluates on the fly whether newly generated candidate features should be committed to the cache and whether existing cached features should be evicted (Sec.~\ref{sec:cache_management}).

\vspace{-0.1cm}
\subsection{Cache Management in SpaceCache}
\label{sec:cache_management}

\begin{wrapfigure}[15]{R}{0.5750\textwidth}
  \vspace{-20pt}
  \centering
  \includegraphics[width=\linewidth]{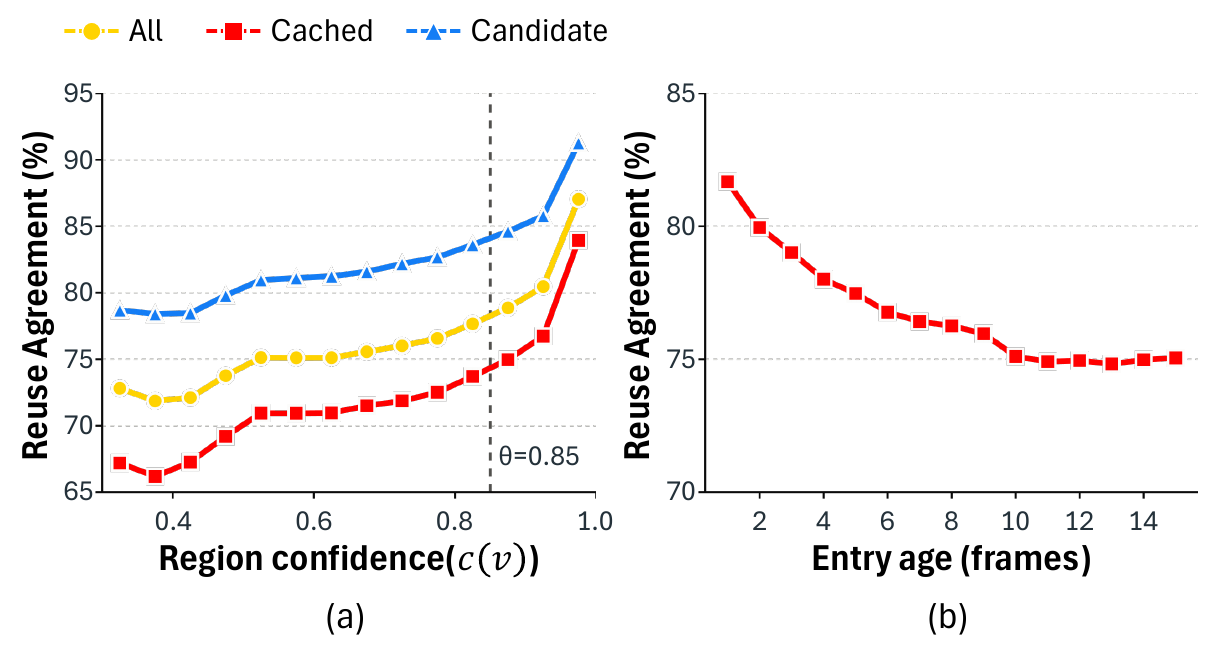}
  \vspace{-20pt}

\caption{\textbf{Reuse Agreement of cache entries} (Stage-3, Waymo with Utonia). (a) By region confidence $c(v)$ at $L{=}8$ for all, cached, and candidate entries; the dashed line marks $\theta{=}0.85$. (b) By entry age with $L{=}\infty$.}
    
  \label{fig:confidence}
  \vspace{-10pt}
\end{wrapfigure}

\vspace{-0.1cm}
\textbf{Confidence filtering---dynamic commit and eviction.} We use decoder prediction confidence as a proxy for feature reliability, filtering out low-confidence features from both new candidates and cached entries. As Figure~\ref{fig:confidence} shows, Reuse Agreement (Sec.~\ref{sec:temporal_locality}) increases with region confidence for both candidate and cached entries, indicating that higher-confidence features tend to remain more semantically consistent across frames.

The Commit Gate computes region confidence from the output logits of the current frame. For each point, we take its maximum class probability, and the confidence of a region $v$ is the average of these probabilities over the input points in $v$:
\begin{equation}
  c(v) = \frac{1}{|\Omega_v|}
  \sum_{j \in \Omega_v} \max_{k}\, \sigma(z_j)_k,
\label{eq:confidence}
\end{equation}
where $\Omega_v$ is the set of input points associated with region $v$, $z_j$ is the logit vector of point $j$, and $\sigma$ is the softmax. With a pre-defined confidence threshold $\theta$, candidate entries with $c(v){\leq}\theta$ are discarded, and cached entries at or below the same threshold are evicted. %

\vspace{-0.1cm}
\textbf{Lifetime--static eviction.} Confidence filtering does not explicitly bound how long a cached feature can persist, while Figure~\ref{fig:confidence} shows that reuse agreement decreases with cache age. In particular, older entries are progressively less likely to remain semantically consistent with the corresponding current-frame features. SpaceCache therefore imposes a maximum lifetime $L$ on cache entries. Each newly committed entry is assigned an expiration age in $\{1,\dots,L\}$ from a hash of its region index, preventing all entries from expiring simultaneously (Appendix~\ref{app:impl_spacecache}). Once expired, the corresponding region is recomputed when observed again. A smaller $L$ refreshes features more frequently at the cost of additional computation. We study this trade-off in Sec.~\ref{sec:ablation}.

\vspace{-0.1cm}
\subsection{Computation Reduction and Overhead}
\label{sec:cost}

\textbf{Computation reduction.} Let $N_l$ be the number of tokens at stage $l$ and $h_l$ the cache-hit rate; after the Mask Gate, the enclosed encoder blocks process only $(1-h_l)N_l$ tokens. The major operations in point-cloud transformer blocks scale linearly with the number of tokens: windowed attention costs $\mathcal{O}(WN_l)$ for a fixed window size $W$, token-wise projections and MLPs cost $\mathcal{O}(N_l)$, and submanifold sparse convolution costs $\mathcal{O}(KN_l)$ for a fixed kernel size $K$. Hence, SpaceCache reduces the computation of these operations approximately in proportion to $h_l$.

\textbf{Overhead.} SpaceCache incurs a one-time per-frame overhead for region-index generation, cache lookup, feature assembly, and cache update. In our setting, all cache operations together take about 5\,ms per frame, while the cache occupies approximately 70\,MiB with single-precision features. The overhead remains small because index generation uses simple coordinate transformation and quantization, lookup performs parallel binary searches over scalar indices, and feature assembly only restores cached and computed features to their token positions. Since these costs are largely independent of the number of enclosed blocks, they are better amortized as more blocks are bypassed. We provide a detailed breakdown in Sec.~\ref{sec:efficiency}.

\textbf{Stage selection for SpaceCache.} Following the design takeaway from Sec.~\ref{sec:motivation}, among the five encoder stages (S0-S4), Stages 3 and 4 exhibit both high temporal overlap and high feature reusability. We therefore primarily apply SpaceCache to these deeper stages. SpaceCache remains applicable to any sequence of consecutive blocks within a stage that preserves token coordinates, and we evaluate different stage selections in Sec.~\ref{sec:ablation}.

\vspace{-0.1cm}
\section{Experimental Results}
\label{sec:experiments}

\vspace{-0.1cm}
\textbf{Dataset and Implementation Details.} We use the full validation sets of two outdoor LiDAR datasets, nuScenes~\citep{caesar2020nuscenes} (6,019 frames, 150 scenes) and Waymo~\citep{sun2020waymo} (5,976 frames, 202 scenes), whose labeled sweeps arrive at 2\,Hz. For the indoor stream, we convert each ScanNet~\citep{dai2017scannet} RGB-D frame with its camera pose into a point-cloud stream subsampled to 3\,Hz, using 10 validation scenes (2,221 frames). We integrate SpaceCache into two pretrained point-cloud networks, Utonia~\citep{utonia} and LitePT-B~\citep{litept}. By default, SpaceCache is applied to encoder stages 3 and 4 (SpaceCache$_{\text{Stage-}3,4}$); we also evaluate a stage-3-only variant (SpaceCache$_{\text{Stage-}3}$). We set the lifetime $L$ to 8/4 frames for outdoor/indoor streams and the confidence threshold $\theta$ to 0.85. Outdoor/indoor latency is measured on an NVIDIA RTX A6000/Jetson AGX Orin. All models use a single forward pass without test-time augmentation. 

\textbf{Baselines.} We compare SpaceCache with token-merging methods that aggregate redundant tokens within each frame: ToMe~\citep{tome} and ALGM~\citep{algm}, designed for 2D image transformers and applied to point tokens, and GitMerge3D~\citep{gitmerge3d} without its adaptive patch partition\footnote{The adaptive version reduces each redundant attention patch from 1024 tokens to 32, which either loses accuracy or runs slower than the version we report (Appendix~\ref{app:impl_gitmerge3d}).}. Implementation details are provided in Appendix~\ref{app:implentation_details}.

\begin{table*}[t]
\centering
\renewcommand{\arraystretch}{1.00}
\setlength{\tabcolsep}{3pt}
\fontsize{7.5pt}{7.5pt}\selectfont

\caption{\textbf{Comparison on outdoor streams.} Each method is shown at its fewest-FLOPs setting within a 1-point mIoU drop from the backbone (rings in Figure~\ref{fig:pareto}). Latency (ms) and GFLOPs are measured per frame on representative scenes with dense attention. Dashes indicate that no setting satisfies the 1-point drop. $^{*}$: 2D method applied to point tokens. $^{\dagger}$: without the adaptive patch partition.}

    \vspace{-0.2cm}
    \label{tab:outdoor}
    \setlength{\tabcolsep}{2.5pt}
    \begin{tabular}{lcccccccc}
        \toprule
        & \multicolumn{4}{c}{Waymo}
        & \multicolumn{4}{c}{nuScenes} \\
        \cmidrule(lr){2-5}
        \cmidrule(lr){6-9}
        Method
        & mIoU (\%)$\uparrow$
        & Latency$\downarrow$
        & GFLOPs$\downarrow$
        & Speedup$\uparrow$
        & mIoU (\%)$\uparrow$
        & Latency$\downarrow$
        & GFLOPs$\downarrow$
        & Speedup$\uparrow$ \\
        \midrule

        \textbf{Utonia}
        & 67.17 & 721 & 4438 & 1.00$\times$
        & 77.29 & 226 & 1269 & 1.00$\times$ \\

        \quad + ToMe$^{*}$
        & 66.35 & 691 & 3794 & 1.04$\times$
        & 76.31 & 255 & 1024 & 0.89$\times$ \\

        \quad + ALGM$^{*}$
        & 66.24 & 633 & 3751 & 1.14$\times$
        & 76.36 & 209 & 1060 & 1.08$\times$ \\

        \quad + GitMerge3D$^{\dagger}$
        & 66.38 & 706 & 3864 & 1.02$\times$
        & 76.32 & 237 & 1123 & 0.95$\times$ \\

        \quad + SpaceCache$\,_{\text{Stage-}3}$
        & 66.25 & 563 & 3177 & 1.28$\times$
        & 76.47 & 192 & 982 & 1.18$\times$ \\

        \quad + SpaceCache$\,_{\text{Stage-}3,4}$
        & 66.50 & \textbf{537} & \textbf{2941} & \textbf{1.34$\times$}
        & 76.38 & \textbf{187} & \textbf{918} & \textbf{1.21$\times$} \\

        \midrule

        \textbf{LitePT-B}
        & 72.82 & 307 & 2392 & 1.00$\times$
        & 80.04 & 98 & 683 & 1.00$\times$ \\

        \quad + ToMe$^{*}$
        & 72.00 & 245 & 1488 & 1.25$\times$
        & 79.33 & 101 & 492 & 0.97$\times$ \\

        \quad + ALGM$^{*}$
        & -- & -- & -- & --
        & -- & -- & -- & -- \\

        \quad + GitMerge3D$^{\dagger}$
        & 72.05 & 287 & 2020 & 1.07$\times$
        & 79.05 & 101 & 605 & 0.97$\times$ \\

        \quad + SpaceCache$\,_{\text{Stage-}3}$
        & 71.86 & 202 & 1374 & 1.52$\times$
        & 79.21 & 76 & 468 & 1.29$\times$ \\

        \quad + SpaceCache$\,_{\text{Stage-}3,4}$
        & 72.11 & \textbf{186} & \textbf{1226} & \textbf{1.65$\times$}
        & 79.14 & \textbf{75} & \textbf{444} & \textbf{1.31$\times$} \\
        \bottomrule
    \end{tabular}
\end{table*}

\begin{figure}[t]
    \centering
    \includegraphics[width=1\textwidth]{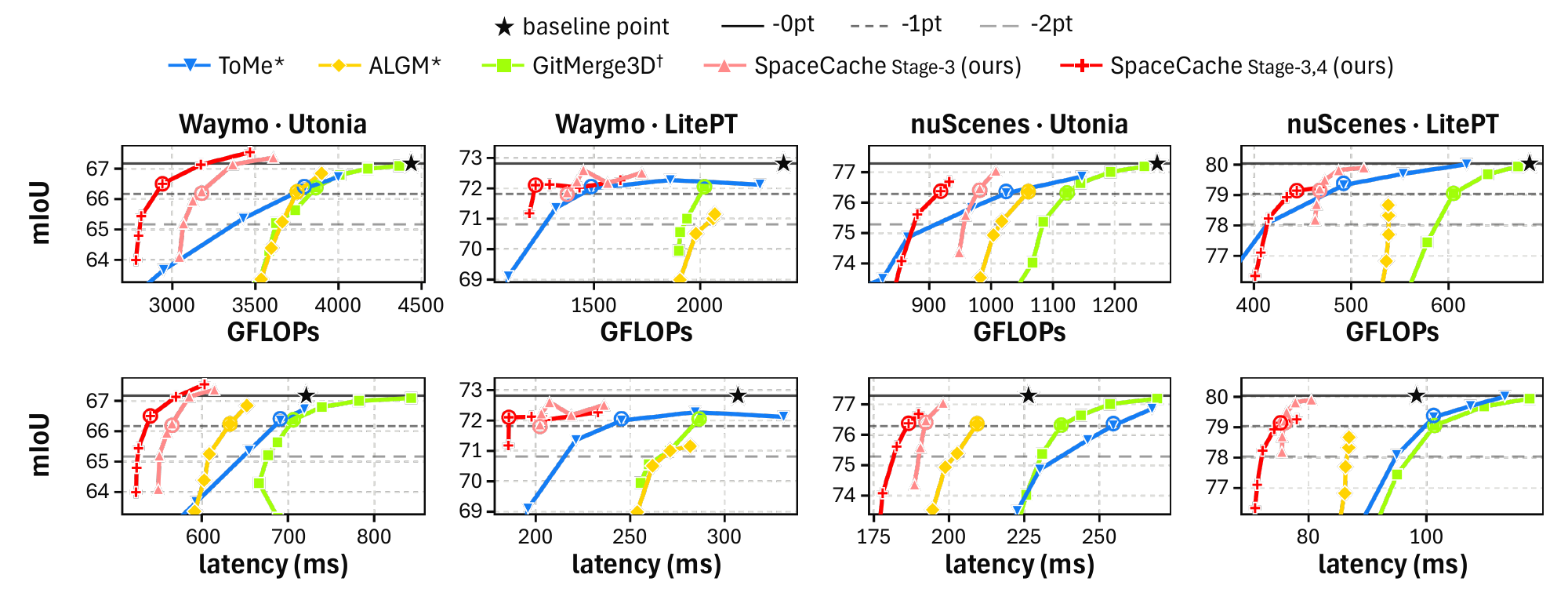}
    \vspace{-0.6cm}
    \caption{\textbf{Accuracy-efficiency trade-off on outdoor streams.} Curves sweep the merge ratio (ToMe, GitMerge3D), similarity threshold (ALGM), or confidence threshold $\theta$ (SpaceCache), plotting mIoU against FLOPs (top) and latency (bottom). Solid line and star--backbone; dashed lines--1- and 2-point mIoU drops;  Rings--fewest-FLOPs point of each method within the 1-point drop.
    }
    \label{fig:pareto}
    \vspace{-0.3cm}
\end{figure}

\begin{table*}[t]
    \centering
    \renewcommand{\arraystretch}{1.00}
    \setlength{\tabcolsep}{3pt}
    \fontsize{7.5pt}{7.5pt}\selectfont
    
    \centering
    \caption{\textbf{Comparison on indoor streams.} Each method is shown at its fewest-FLOPs setting within 1-point and 2-point mIoU drops. Kernel time (ms) is averaged over 50 representative frames. The notation follows Table~\ref{tab:outdoor}.}
    \vspace{-0.2cm}
    \label{tab:indoor}
    \setlength{\tabcolsep}{2.5pt}
    \begin{tabular}{lcccccccc}
        \toprule
        & \multicolumn{4}{c}{mIoU drop $\leq$ 1.0 pt}
        & \multicolumn{4}{c}{mIoU drop $\leq$ 2.0 pt} \\
        \cmidrule(lr){2-5}
        \cmidrule(lr){6-9}
        Method
        & mIoU (\%)$\uparrow$ & Kernel time$\downarrow$ & GFLOPs$\downarrow$ & Speedup$\uparrow$
        & mIoU (\%)$\uparrow$ & Kernel time$\downarrow$ & GFLOPs$\downarrow$ & Speedup$\uparrow$ \\
        \midrule
        \textbf{LitePT-B}
        & 50.63 & 43.3 & 39.9 & 1.00$\times$
        & 50.63 & 43.3 & 39.9 & 1.00$\times$ \\
        \quad + ToMe$^{*}$
        & 49.72 & 41.1 & \textbf{28.0} & 1.06$\times$
        & 48.80 & 40.5 & 26.6 & 1.07$\times$ \\
        \quad + ALGM$^{*}$
        & 49.80 & 39.9 & 32.1 & 1.09$\times$
        & 48.75 & 37.2 & 29.7 & 1.16$\times$ \\
        \quad + GitMerge3D$^{\dagger}$
        & 49.64 & 47.2 & 37.3 & 0.92$\times$
        & 48.65 & 46.8 & 37.1 & 0.93$\times$ \\
        \quad + SpaceCache$\,_{\text{Stage-}3}$
        & 49.81 & \textbf{36.8} & \textbf{28.0} & \textbf{1.18$\times$}
        & 48.68 & \textbf{35.5} & \textbf{25.6} & \textbf{1.22$\times$} \\
        \quad + SpaceCache$\,_{\text{Stage-}3,4}$
        & 49.68 & 38.8 & 31.0 & 1.12$\times$
        & 48.80 & 36.2 & 26.2 & 1.20$\times$ \\
        \bottomrule
    \end{tabular}
\end{table*}

\begin{table*}[t]
    \centering
    \caption{\textbf{Ablation study on Waymo with Utonia.} (a) Lifetime control. (b) Stage selection. Gray: default setting; $L{=}1$ reproduces the backbone. Hit: cache-hit fraction of each cached stage. $\Delta$Dyn./$\Delta$Sta.: accuracy change (points) on dynamic/static regions at stage-3 relative to $L{=}1$ (83.7/94.8\%; Appendix~\ref{app:metrics}). $\Delta$FLOPs: encoder FLOPs change relative to the backbone.}
 
    \vspace{-0.2cm}
    \label{tab:ablation}
    \scriptsize
    \setlength{\tabcolsep}{3pt}
    \begin{subtable}[t]{0.60\textwidth}
        \centering
        \caption{Cache management (commit and eviction)}
        \vspace{-0.1cm}
        \label{tab:ablation_lifetime}
        \begin{tabular}{ccccccccc}
            \toprule
            & \multicolumn{4}{c}{Age only} & \multicolumn{4}{c}{Age + Confidence} \\
            \cmidrule(lr){2-5}
            \cmidrule(lr){6-9}
            $L$ & Speedup$\uparrow$ & mIoU$\uparrow$ & $\Delta$\,Dyn. & $\Delta$\,Sta. & Speedup$\uparrow$ & mIoU$\uparrow$ & $\Delta$\,Dyn. & $\Delta$\,Sta. \\
            \midrule
            1  & 1.00$\times$ & 67.18 & --     & --     & 1.00$\times$ & 67.18 & --     & --     \\
            2  & 1.09$\times$ & 67.14 & $-$0.5 & $-$0.1 & 1.10$\times$ & 67.12 & $-$0.5 & 0.0    \\
            4  & 1.24$\times$ & 65.84 & $-$1.9 & $-$0.3 & 1.21$\times$ & 67.11 & $-$1.1 & 0.0    \\
            8  & 1.35$\times$ & 63.99 & $-$3.0 & $-$0.8 & \cellcolor{gray!15}1.25$\times$ & \cellcolor{gray!15}66.91 & \cellcolor{gray!15}$-$1.8 & \cellcolor{gray!15}0.0 \\
            16 & 1.44$\times$ & 62.79 & $-$4.1 & $-$1.0 & 1.33$\times$ & 66.68 & $-$3.0 & 0.0    \\
            32 & 1.44$\times$ & 62.37 & $-$5.3 & $-$0.9 & 1.37$\times$ & 66.53 & $-$3.9 & $-$0.1 \\
            \bottomrule
        \end{tabular}
    \end{subtable}%
    \hfill
    \begin{subtable}[t]{0.37\textwidth}
        \centering
        \caption{Stage selection for SpaceCache}
        \vspace{-0.1cm}
        \label{tab:ablation_placement}
        \begin{tabular}{ccccc}
            \toprule
            Stage & \makecell{Hit\\(\%)} & \makecell{$\Delta$FLOPs\\(\%)} & Speedup$\uparrow$ & mIoU$\uparrow$ \\
            \midrule
            0   & 20      & $-$0.7  & 1.00$\times$ & 67.00 \\
            1   & 32      & $-$2.1  & 1.01$\times$ & 65.28 \\
            2   & 42      & $-$5.0  & 1.03$\times$ & 65.42 \\
            3   & 48      & $-$34.7 & 1.21$\times$ & 66.60 \\
            4   & 52      & $-$6.5  & 1.03$\times$ & 67.47 \\
            \midrule
            \rowcolor{gray!15}
            3+4 & 48 / 51 & $-$41.1 & 1.25$\times$ & 66.91 \\
            \bottomrule
        \end{tabular}
    \end{subtable}

\vspace{-0.3cm}
\end{table*}

\subsection{Comparison with Prior Methods}
\label{sec:outdoor}

\textbf{Comparison in outdoor scenario.} Table~\ref{tab:outdoor} compares SpaceCache with ToMe, ALGM, and GitMerge3D on Waymo and nuScenes with Utonia and LitePT-B, each at its  fewest-FLOPs setting within a 1-point mIoU drop. SpaceCache is the fastest in all four settings, reaching up to 1.65$\times$ on Waymo with LitePT-B (307$\rightarrow$186\,ms) by removing 49\% of FLOPs. 

Figure~\ref{fig:pareto} extends this comparison over the full sweep of each method. Within the 2.0-point drop, SpaceCache achieves a better trade-off than token merging in all four settings, in both FLOPs (top) and latency (bottom). Two properties underlie this advantage: more FLOPs removed and better FLOPs-to-latency conversion. We analyze both in Sec.~\ref{sec:efficiency}.

\textbf{Comparison in indoor scenario.} Table~\ref{tab:indoor} reports each method using LitePT-B on ScanNet at its setting with the fewest FLOPs within 1-, 2-point mIoU drops. Since host-side launch overhead dominates the latency of small indoor frames, we report the average GPU kernel time over 50 representative frames. SpaceCache$_{\text{Stage-}3}$ is the fastest with the fewest FLOPs at both drops, reaching 1.22$\times$ with 36\% fewer GFLOPs within the 2-point drop. The speedup is smaller than outdoors because the denser indoor points are merged more aggressively by downsampling.

\suppressfloats[t]

\subsection{Ablation Study}
\label{sec:ablation}

\textbf{Cache management.} Table~\ref{tab:ablation}(a) examines SpaceCache's two control knobs, which trade accuracy for efficiency rather than having a single optimal value: the lifetime $L$ and confidence filtering with threshold $\theta$. With confidence filtering (``Age + Confidence''), increasing $L$ from 1 to 32 raises the speedup to 1.37$\times$ but lowers the mIoU by 0.65 points, as older features survive longer in the cache; the drop is more pronounced on dynamic objects ($\Delta$\,Dyn.\ of $-$3.9). Without filtering (``Age only''), the mIoU drops far more steeply (4.81 points at $L{=}32$), and even at a similar speedup it trails the filtered setting (65.84\% at 1.24$\times$ for $L{=}4$ vs.\ 66.91\% at 1.25$\times$ for $L{=}8$).

\textbf{Stage selection.} %
Table~\ref{tab:ablation}(b) shows how performance changes with the stage which SpaceCache is applied to. The speedup grows with the number of enclosed blocks, peaking at stage-3, which holds 12 of the 24 blocks. The accuracy loss shrinks with depth from stage-1 to 4, consistent with the semantic consistency in Figure~\ref{fig:motivation}; despite its low consistency, stage-0 loses little, as it reuses the fewest tokens and all later stages refine them. Stage-3,\,4 together thus give the largest speedup at a small accuracy drop, which we use as the default.

\vspace{-0.3cm}
\begin{figure}[t]
    \centering
    \includegraphics[width=0.95\textwidth]{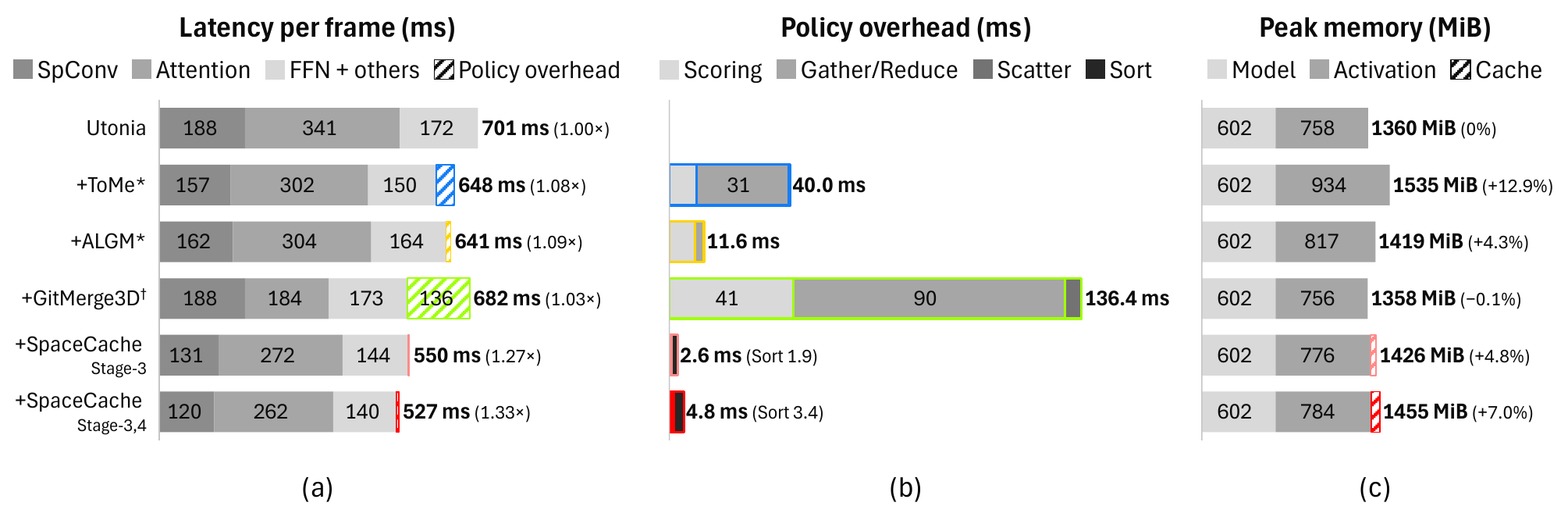}
    \vspace{-0.4cm}
    \caption{
    \textbf{Latency and memory breakdown} on the representative scenes of Waymo with Utonia, each method within a 2.0-point mIoU drop.
    (a) Latency per frame, split into the backbone layers and the policy overhead.
    (b) Policy overhead per frame, split into its components.
    (c) Peak memory with flash-attention, split into the model weights, the activation memory, and the cache.
    }
    \label{fig:breakdown}
    \vspace{-0.4cm}    
\end{figure}

\subsection{Efficiency Analysis}
\label{sec:efficiency}

\textbf{Where the gain comes from.} SpaceCache outperforms token merging at similar accuracy (Figure~\ref{fig:pareto}) for two reasons, decomposed on Waymo with Utonia within a 2-point drop. (1)~\textit{More FLOPs removed.} Cache-hit tokens skip every block of a stage at a negligible scoring cost, so SpaceCache shrinks the sparse convolution, attention, and FFN together, removing 37\% of the total FLOPs (Appendix~\ref{app:breakdown_details}). GitMerge3D instead merges tokens only within attention, and its per-block scoring leaves a net reduction of only 18\%. (2)~\textit{Better FLOPs-to-latency conversion.} SpaceCache runs its cache operations once per stage, costing only 4.8\,ms, mostly for sorting at the Commit Gate. Merging repeats scoring at every block and accumulates merged features through memory-bound gather/reduce, so GitMerge3D adds 136\,ms of overhead (Figure~\ref{fig:breakdown}). Even ToMe, the merging method that removes the most FLOPs here (23\%; Appendix~\ref{app:breakdown_details}), reduces the latency by only 7\%, whereas SpaceCache$_{\text{Stage-}3,4}$ turns a 37\% FLOPs reduction into a 25\% latency reduction (Figure~\ref{fig:breakdown}(a)).

\textbf{Memory breakdown and overhead.} The memory overhead of SpaceCache is also small. As shown in Figure~\ref{fig:breakdown}(c), SpaceCache raises the peak memory by only 95\,MiB (7.0\%): 69\,MiB for the cache itself and 26\,MiB of activations for holding cache-miss features until the Commit Gate. %

\section{Related Work}
\label{sec:related}

\vspace{-0.1cm}
\textbf{Efficient point-cloud backbones.} Point-cloud backbones have moved from sparse convolution~\citep{graham2018submanifold, choy2019minkowski} to point transformers~\citep{zhao2021pt, wu2022ptv2}, and many designs make them faster by sampling points randomly~\citep{hu2020randlanet}, grouping points into superpoints~\citep{robert2023spt}, or simplifying attention~\citep{wu2024ptv3, litept}. Such designs, however, need training, which is costly now that backbones are pretrained at scale~\citep{wu2025sonata, utonia}. We instead speed up existing pretrained backbones on point-cloud streams without training.

\textbf{Token reduction in vision.} A common way to speed up transformer networks in vision is to process fewer tokens. Token merging combines similar tokens~\citep{tome, liang2022expedit, algm, gitmerge3d}, pruning removes unimportant ones~\citep{liang2022evit, chen2023sparsevit, zhao2023ada3d, baiocchi2025adaptive, li2025gsp}, and halting stops tokens at intermediate layers in 2D~\citep{yin2022avit, tang2023dtop} and 3D~\citep{ye2023efficient}. Some point networks instead group points into fewer tokens~\citep{xu2021yogo, wang2025satformer}. These methods require training by design, except for merging. Merging, however, often searches for similar tokens again at every block and accumulates several tokens into one through slow memory accesses, which limits its speed-up~\citep{toma}. SpaceCache instead selects tokens cheaply and only once per stage, and copies each cached feature one-to-one without accumulating tokens.

\textbf{Temporal reuse in point-cloud streams.} While many methods use past frames for accuracy~\citep{liu2019meteornet, cao2020asapnet, li2023memoryseg, he2023lef, dang2025faster}, few use them to save computation. Of these, TempNet~\citep{zhou2021tempnet} propagates key-frame features with trained modules, and SSCATeR~\citep{dow2026sscater} recomputes convolutions only where points changed. Closest to our work, EVE~\citep{peng2021eve} prunes input points near the previous frame and trains a module to recover lost context. SpaceCache instead keeps all input points, reuses only deep-stage features, and needs no training.

\section{Conclusion}
\vspace{-0.1cm}
We presented SpaceCache, a training-free intermediate-feature-reuse method for efficient point-cloud stream processing. By exploiting the high temporal overlap and feature reusability of deeper encoder stages, SpaceCache skips redundant computation while preserving dense token representations. Across outdoor and indoor semantic-segmentation benchmarks, SpaceCache consistently improves the accuracy-efficiency trade-off over existing token-reduction methods, with low runtime and memory overhead.


\bibliography{iclr2027_conference}
\bibliographystyle{iclr2027_conference}

\appendix

\clearpage
\section{Token Regions and Reuse Agreement}
\label{app:region_ra}

\subsection{Token Regions}
\label{app:region}

\textbf{Pooling chain.} Each downsampling layer merges the tokens that fall into the same voxel of the next stage into one token and records which token each of them went to, $\pi_l(i)$ for a token $i$ of stage $l{-}1$. We call the chain of these records the \emph{pooling chain}. Following it from the input points, the tokens of stage 0, every input point $j$ reaches exactly one token of stage $l$, so each token $v$, which represents one spatial region, covers the input points
\begin{equation}
\Omega_v = \{\, j : \pi_l(\pi_{l-1}(\cdots \pi_1(j))) = v \,\}.
\end{equation}

\textbf{Token labels.} The label of a token is the majority ground-truth label of its input points, $y_v = \arg\max_c |\{ j \in \Omega_v : y_j = c \}|$, ignoring the ignore label, and a token is \emph{pure} if all of them carry the same label. The following sections use the same pooling chain and labels.

\textbf{Token accuracy.} Our accuracy measures score a group of tokens by the input points they cover through the pooling chain. For a group $G_t$ of tokens in a frame $t$, the token accuracy is
\begin{equation}
\mathrm{Acc}(G_t) = \frac{\sum_{v \in G_t} \sum_{j \in \Omega_v} \mathbf{1}[\hat{y}_j = y_j]}{\sum_{v \in G_t} |\Omega_v|},
\end{equation}
where $\hat{y}_j$ is the predicted class, the argmax of the logits, and $y_j$ is the ground-truth label of the voxel of point $j$; points with the ignore label are excluded. It is the recall of the points covered by the group within the frame, with each token weighted by its covered-point count. We then average $\mathrm{Acc}(G_t)$ over all frames, weighted by the covered-point count of each frame. Appendix~\ref{app:metrics} uses this measure with different groups $G_t$.

\subsection{Reuse Agreement}
\label{app:reuse_agreement}

Reuse Agreement (Figures~\ref{fig:motivation}(c) and~\ref{fig:confidence}) measures whether a reused feature predicts the same class as the feature the backbone computes for the same region in the current frame.

\textbf{Class prototypes.} At stage $l$, the prototype of class $c$ is the mean current-frame feature over $F_c^l$, the pure tokens of class $c$:
\begin{equation}
\mu_c^l = \frac{1}{|F_c^l|} \sum_{f \in F_c^l} f .
\end{equation}
Only classes with at least 20 such tokens in the frame get a prototype.

\textbf{Agreement.} Each feature $f$ is mapped to a probability distribution over the classes by the softmax of its cosine similarities to the prototypes,
\begin{equation}
q(f) = \mathrm{softmax}\big([\cos(f,\mu_1^l), \ldots, \cos(f,\mu_C^l)] \,/\, T_l\big),
\end{equation}
where the temperature $T_l$ is calibrated per stage so that $q$ has the same mean normalized entropy of 0.5 at every stage, for a fair comparison between stages. The Reuse Agreement of a token is the overlapping area of the distributions of its current feature $f^t$ and its reused feature $\tilde{f}$ over the classes:
\begin{equation}
\mathrm{RA} = \sum_c \min\big(q_c(f^t),\, q_c(\tilde{f})\big) \in [0, 1].
\end{equation}
RA is 1 when $q(f^t)$ and $q(\tilde{f})$ are identical and 0 when they share no class.

For Figure~\ref{fig:motivation}(c), the reused feature is the one computed for the same region in frame $t{-}1$. We average RA over all overlapped tokens of 60 consecutive frame pairs per dataset and plot the mean over the three datasets for each backbone. For Figure~\ref{fig:confidence}, the reused feature is a cached or candidate feature.

\clearpage
\section{Implementation Details}
\label{app:implentation_details}

This section describes how SpaceCache and the three baselines are implemented. All methods are implemented in the Pointcept codebase~\citep{pointcept2023}\footnote{\url{https://github.com/Pointcept/Pointcept}} and use the same pretrained weights without training, and each baseline is swept over its main parameter (Figure~\ref{fig:pareto}; Table~\ref{tab:swept} lists the swept values). On LitePT-B~\citep{litept}, whose attention layers are only in the last two encoder stages, ToMe and GitMerge3D merge in those blocks.

All methods run with batch size 1 and dense attention instead of flash-attention~\citep{dao2022flashattention}. Note that SpaceCache itself is compatible with flash-attention, while GitMerge3D adds to the attention scores a bias for the number of tokens merged into each key, which flash-attention cannot express. The attention is computed in bfloat16 for Utonia~\citep{utonia} and float16 for LitePT-B, as in each backbone, and all other layers in float32.

\subsection{SpaceCache}
\label{app:impl_spacecache}

\textbf{Settings.} The region of each stage is its own voxel, $s_l = 2^l g$, with $g = 5$\,cm on Waymo~\citep{sun2020waymo} and nuScenes~\citep{caesar2020nuscenes} and 2\,cm on ScanNet~\citep{dai2017scannet}. Each cached stage keeps its own SpaceCache. When several tokens of a frame fall into the same region, their features are averaged into one entry. The cache is emptied at the start of every sequence. The lookup, the feature assembly, and the cache update run as fused Triton~\citep{tillet2019triton} kernels on the RTX A6000 and in PyTorch on the Jetson AGX Orin.

\textbf{Region key.}\label{app:key_mapping} $\phi$ shifts the three indices of a region $v=(v_x,v_y,v_z)$ to non-negative values and packs them into separate bit fields of one 64-bit integer, 24 bits for $v_x$ and $v_y$ and 15 bits for $v_z$:
\begin{equation}
\phi(v) = \big((v_x + 2^{23}) \ll 39\big) \,|\, \big((v_y + 2^{23}) \ll 15\big) \,|\, (v_z + 2^{14}),
\end{equation}
where $\ll$ is the left bit shift and $|$ the bitwise OR. It costs three additions, two shifts, and two ORs per token, and it is one-to-one for $|v_x|, |v_y| < 2^{23}$ and $|v_z| < 2^{14}$, \textit{i.e.}, within $\pm 2^{23} s_l$ horizontally and $\pm 2^{14} s_l$ vertically of the global origin. The range doubles with each stage:
\begin{center}
\small
\begin{tabular}{lcccc}
\toprule
Dataset & Stage & Cell $s_l$ & Horizontal & Vertical \\
\midrule
Waymo, nuScenes & 3 & 0.4\,m & $\pm$3,355\,km & $\pm$6.5\,km \\
ScanNet & 3 & 0.16\,m & $\pm$1,342\,km & $\pm$2.6\,km \\
Waymo (placement ablation) & 0 & 5\,cm & $\pm$419\,km & $\pm$819\,m \\
\bottomrule
\end{tabular}
\end{center}
All are far larger than any sequence in the benchmarks.

\textbf{Lifetime.} The Age of an entry is implemented as a due frame rather than a counter. Each region index $v$ has a fixed phase, the linear hash $p_1 v_x + p_2 v_y + p_3 v_z$ with large primes $p_i$, and an entry committed at frame $t$ is due at the next frame congruent to that phase modulo $L$: $d_v = t + 1 + (\mathrm{phase}(v) - (t{+}1)) \bmod L$. The lookup counts an entry as a cache hit only while $t < d_v$. Since the due frame is modular in the region index instead of counted from the insertion frame, every region is refreshed at a fixed $L$-frame cadence and roughly one $L$-th of the resident entries expires per frame, which keeps the cache-hit ratio stable across frames.

\textbf{Regions without training labels.} Waymo labels only its top LiDAR, so the preprocessing gives the ignore label to every point of its four short-range LiDARs around the vehicle. nuScenes labels the points that hit its own car as \textit{vehicle.ego}, which the benchmark ignores. These points receive no training loss, so the Commit Gate does not insert a region that contains them. We take these points from the ignore label, which also covers the few points of classes that the benchmark ignores. At inference, the same points can be identified without labels, as the points of the short-range LiDARs on Waymo and the points inside a fixed box around the ego vehicle on nuScenes. The ignored points never count toward the mIoU, and we do not apply this exclusion on ScanNet.

\subsection{ToMe}
\label{app:impl_tome}

ToMe~\citep{tome}\footnote{\url{https://github.com/facebookresearch/ToMe}} merges tokens by bipartite matching between the attention and the MLP of every attention block. Within each serialized attention patch, the tokens are split alternately into two sets, each token of the first set is matched to its most similar token of the second set by the cosine similarity of their attention keys, and the best-matched pairs are averaged into one token, without proportional attention. The sweep parameter $r$ is the merging ratio, the fraction of the tokens of each patch merged in a block. A token merged at stage $l$ stays merged until the decoder unpooling from stage $l$ to $l{-}1$, which copies its feature back to its original tokens. Stage 0 has no unpooling below it, so its merges are unmerged at the end of the decoder, before the segmentation head.

\subsection{ALGM}
\label{app:impl_algm}

ALGM~\citep{algm}\footnote{\url{https://github.com/tue-mps/algm-segmenter}} merges twice. Local merging (CLAP), after the attention of the first block of stage 0, merges each $4{\times}4{\times}4$-voxel window into one token if the mean pairwise cosine similarity of its tokens exceeds $\tau$. The windows are offset by two voxels from the downsampling grid so that the following downsamplings do not absorb the merge. Global merging (GBM), after the attention of the first block of stage 3, splits all tokens of the stage alternately into two sets and matches them by bipartite matching. Following the official implementation, it keeps a pair only if its similarity exceeds $\mu + \sigma / b$, where $\mu$ and $\sigma$ are the mean and standard deviation of the best-match similarities and $b$ is the block index, so the number of merges adapts to each frame. Both merges stay until the decoder returns to their stage and copies the merged features back: GBM at the first decoder block of stage 3, and CLAP at the first decoder block of stage 0. On LitePT-B, which has no attention in stage 0 and no decoder blocks, CLAP is applied after the first block of stage 0, and both merges are copied back at the corresponding unpooling.

\subsection{GitMerge3D}
\label{app:impl_gitmerge3d}

GitMerge3D~\citep{gitmerge3d}\footnote{\url{https://github.com/anhtuanhsgs/GitMerge3D}} merges inside the attention of every attention block. 
Each serialized attention patch is divided into bins of $\lceil 1/(1-r) \rceil$ tokens, and one random token of each bin becomes a destination.
A fraction $r$ of the tokens in each patch are averaged into their most similar destinations, based on the cosine similarity of the value vectors in each attention head.
The attention then runs on the merged queries, keys, and values, and its output is copied back to the original tokens right away.

We report GitMerge3D without the adaptive patch partition ($^{\dagger}$ in the tables). The adaptive version splits the patches by a global energy score. Patches with a high energy are merged at ratio $r$ as above, and the other patches are reduced from 1024 to 32 tokens each. At the same merge ratio, reducing a 1024-token patch to 32 tokens lowers the accuracy considerably on Waymo. On nuScenes and the indoor stream, the accuracy is preserved, but computing the energy and splitting the patches offset the savings from merging (Table~\ref{tab:gitmerge3d_adaptive}).

\begin{table}[h]
\centering
\footnotesize
\setlength{\tabcolsep}{4pt}
\caption{\textbf{GitMerge3D with and without its adaptive patch partition} at the same merge ratio $r{=}0.5$. Speedup is measured in latency on Waymo and nuScenes and in kernel time on ScanNet. $^{\dagger}$: without the adaptive patch partition.}
\label{tab:gitmerge3d_adaptive}
\begin{tabular}{lcccccc}
\toprule
& \multicolumn{2}{c}{Waymo} & \multicolumn{2}{c}{nuScenes} & \multicolumn{2}{c}{ScanNet} \\
\cmidrule(lr){2-3}\cmidrule(lr){4-5}\cmidrule(lr){6-7}
Method & mIoU (\%)$\uparrow$ & Speedup$\uparrow$ & mIoU (\%)$\uparrow$ & Speedup$\uparrow$ & mIoU (\%)$\uparrow$ & Speedup$\uparrow$ \\
\midrule
\textbf{Utonia} & 67.17 & 1.00$\times$ & 77.29 & 1.00$\times$ & -- & -- \\
\quad + GitMerge3D$^{\dagger}$ ($r{=}0.5$) & 65.64 & 1.05$\times$ & 74.03 & 1.00$\times$ & -- & -- \\
\quad + GitMerge3D ($r{=}0.5$) & 63.64 & 0.89$\times$ & 73.35 & 0.86$\times$ & -- & -- \\
\midrule
\textbf{LitePT-B} & 72.82 & 1.00$\times$ & 80.04 & 1.00$\times$ & 50.63 & 1.00$\times$ \\
\quad + GitMerge3D$^{\dagger}$ ($r{=}0.5$) & 72.05 & 1.07$\times$ & 74.92 & 1.06$\times$ & 49.44 & 0.92$\times$ \\
\quad + GitMerge3D ($r{=}0.5$) & 55.37 & 1.08$\times$ & 74.87 & 1.07$\times$ & 49.48 & 0.92$\times$ \\
\bottomrule
\end{tabular}
\end{table}

\begin{table}[b!]
\centering
\fontsize{8}{9.5}\selectfont
\setlength{\tabcolsep}{2pt}
\caption{\textbf{Swept parameter values.} The outdoor columns list every point of the Figure~\ref{fig:pareto} curves; the ScanNet column lists the two reported operating points of Table~\ref{tab:indoor}. \emph{off} disables the confidence gate. \textbf{Bold} marks the point picked within a 1.0-point mIoU drop.}
\label{tab:swept}
\begin{tabular}{@{}>{\raggedright\arraybackslash}p{1.6cm}c>{\raggedright\arraybackslash}p{2.05cm}>{\raggedright\arraybackslash}p{2.05cm}>{\raggedright\arraybackslash}p{2.05cm}>{\raggedright\arraybackslash}p{2.05cm}>{\raggedright\arraybackslash}p{2.4cm}@{}}
\toprule
& & \multicolumn{2}{c}{Waymo} & \multicolumn{2}{c}{nuScenes} & ScanNet \\
\cmidrule(lr){3-4} \cmidrule(lr){5-6} \cmidrule(lr){7-7}
Method & Param. & Utonia & LitePT-B & Utonia & LitePT-B & LitePT-B \\
\midrule
ToMe$^{*}$ & $r$ & 0.01, \textbf{0.015}, 0.025, 0.04, 0.045 & 0.01, 0.05, \textbf{0.095}, 0.12, 0.16, 0.2 & 0.01, \textbf{0.02}, 0.025, 0.035, 0.045, 0.05 & 0.02, 0.04, \textbf{0.06}, 0.09, 0.12 & \textbf{0.11}, 0.13 \\
ALGM$^{*}$ & $\tau$ & 0.96, 0.97, 0.98, \textbf{0.995}, 1 & 0.6, 0.7, 0.75, 0.85, 0.95, 0.97 & 0.7, 0.95, 0.97, \textbf{0.995} & 0.9, 0.925, 0.95, 0.98, 1 & \textbf{0.9}, 0.8125 \\
GitMerge3D$^{\dagger}$ & $r$ & 0.1, 0.2, 0.3, \textbf{0.4}, 0.5, 0.6, 0.65, 0.72 & \textbf{0.5}, 0.645, 0.7, 0.72 & 0.1, 0.2, 0.3, \textbf{0.35}, 0.45, 0.5, 0.6 & 0.1, 0.2, \textbf{0.33}, 0.45, 0.6 & \textbf{0.55}, 0.645 \\
SpaceCache\newline Stage-3 & $\theta$ & \emph{off}, 0.55, 0.7, \textbf{0.8}, 0.95, 0.99 & \textbf{\emph{off}}, 0.7, 0.8, 0.95, 0.99 & 0.8, 0.9, \textbf{0.97}, 0.99 & \emph{off}, 0.5, \textbf{0.65}, 0.8, 0.95, 0.99 & \textbf{0.85}, \emph{off} \\
SpaceCache\newline Stage-3,4 & $\theta$ & \emph{off}, 0.5, 0.55, \textbf{0.8}, 0.95, 0.99 & \emph{off}, \textbf{0.6}, 0.8, 0.95, 0.99 & 0.7, 0.85, 0.95, \textbf{0.985}, 0.99 & 0.5, 0.7, 0.825, 0.95, \textbf{0.97}, 0.99 & \textbf{0.99}, 0.8 \\
\bottomrule
\end{tabular}
\end{table}

\subsection{Swept Settings}
\label{app:sweep}

Each method is swept over one parameter: the merge ratio $r$ for ToMe and GitMerge3D, the similarity threshold $\tau$ of the local merge for ALGM, and the confidence threshold $\theta$ for SpaceCache. SpaceCache runs at $L{=}8$ on the outdoor streams and $L{=}4$ on the indoor stream. Table~\ref{tab:swept} lists the swept values.

\clearpage
\section{Dataset and Experimental Setup}
\label{app:setup}

\subsection{ScanNet Video Stream}
\label{app:scannet_video}

\textbf{Frames.} We keep every tenth frame (3\,Hz) of the 30\,Hz RGB-D videos of 10 validation scenes, 2,221 frames in total, and drop the few frames without a valid camera pose. Each depth pixel closer than 8\,m is back-projected with the depth intrinsics and takes the color of the nearest color pixel. The scenes are scene0025\_00, scene0050\_00, scene0169\_00, scene0217\_00, scene0426\_00, scene0488\_00, scene0565\_00, scene0568\_00, scene0580\_00, and scene0583\_00.

\textbf{Labels.} Each point takes the label of its pixel in the 2D label images of ScanNet~\citep{dai2017scannet}, mapped to the 20 benchmark classes as for the fused scenes. All other labels are ignored.

\textbf{Coordinates and normals.} The backbone is pretrained on the fused ScanNet scenes, so we express each frame in the coordinates of its scene with the camera pose. All frames of a sequence thus share one voxel grid. LitePT-B also takes surface normals as input and is pretrained on the normals of the scene meshes, so each point takes the normal of its nearest mesh vertex instead of a normal estimated from the single depth frame.

\subsection{Efficiency Measurement}
\label{app:measurement}

Accuracy uses the full validation sets. Latency, FLOPs, and memory use representative scenes, whole sequences whose per-frame distribution of voxel counts matches the full set: 10 segments (299 frames) of Waymo, 8 scenes (321 frames) of nuScenes, and 2 scenes (395 frames) of ScanNet. On ScanNet, the FLOPs are counted on all 10 scenes.

\textbf{Latency.} For the outdoor scenes we measure on an RTX A6000 and report the mean frame latency. For the indoor scenes we measure on a Jetson AGX Orin and report the mean GPU kernel time per frame over 50 frames, every eighth frame of the scenes, summed over all GPU kernels of each frame. The scenes are streamed in full, so SpaceCache reaches these frames with its true cache state.

\textbf{FLOPs.} The FLOPs count the linear layers, the sparse convolutions with their actual input-output pairs, the attention, and the similarity computations of the merging methods, with one multiply-accumulate as two FLOPs. Normalizations, activations, softmax, and gathers are not counted.

\subsection{Metrics of the Ablation Study}
\label{app:metrics}

In Table~\ref{tab:ablation}, the cache-hit fraction is averaged over all frames except the first of each sequence, where the cache is empty. $\Delta$Dyn. and $\Delta$Static group the third-stage tokens by whether the ground-truth label of their region (Appendix~\ref{app:region}) has changed since the region was last seen. A token is \emph{dynamic} if the label has changed and either the old or the new label is a movable class (car, truck, bus, other vehicle, motorcyclist, bicyclist, pedestrian, bicycle, or motorcycle). A token is \emph{static} if the label has not changed. For each group, we report the change in point accuracy from the backbone ($L{=}1$), where point accuracy is the fraction of points predicted correctly.

\clearpage
\section{Breakdown Details}
\label{app:breakdown_details}
\label{app:overhead}

Figure~\ref{fig:breakdown} compares every method at its point within a 2.0-point mIoU drop on the representative scenes of Waymo with Utonia. This section explains where each method removes computation, where it spends its own overhead, and what it adds to memory (Table~\ref{tab:breakdown}). FLOPs and latency use dense attention, and memory uses flash-attention.

\subsection{FLOPs Saving and Scoring Overhead}
\label{app:breakdown_flops}

Table~\ref{tab:breakdown} shows how many FLOPs each method removes from the network layers and how many its scoring, the step that decides which tokens to reduce, adds back. SpaceCache removes the most and adds the least. As in Section~\ref{sec:cost}, $N_l$ is the number of tokens at stage $l$, $W$ the attention window size, $K$ the kernel size of the sparse convolution, and $h_l$ the cache-hit rate. $D$ is the number of channels, and $r$ is the merge ratio of ToMe and GitMerge3D.

\textbf{FLOPs saving.} A block at stage $l$ costs $\mathcal{O}(WDN_l + KD^2N_l)$, where the first term comes from the attention matrix products $QK^{\top}$ and $AV$ and the second from the linear layers and the sparse convolution. SpaceCache removes $\mathcal{O}\bigl(h_l(WDN_l + KD^2N_l)\bigr)$ at every block of a cached stage. ToMe merges a fraction $r$ of the tokens at every block and thus removes $\mathcal{O}\bigl((1-(1-r)^k)(WDN_l + KD^2N_l)\bigr)$ at its $k$-th block. ALGM merges twice, locally at stage~0 and globally at stage~3. Each merging removes a fraction $\rho_l$ of the tokens, set by the similarities in each frame, and thus $\mathcal{O}\bigl(\rho_l(WDN_l + KD^2N_l)\bigr)$ at every following block. GitMerge3D merges the tokens only between the projections, so it removes $\mathcal{O}\bigl((1-(1-r)^2)WDN_l\bigr)$ from the first term at every block.

\textbf{Scoring overhead.} SpaceCache adds only $\mathcal{O}(N_l)$ ($9N_l$ multiply-adds) once per stage, since it rotates the 3D coordinate of each token and never touches the $D$-dimensional features. The merging methods compare the features of the tokens instead. GitMerge3D adds $\mathcal{O}(WDN_l)$ at every block, the same order as the attention FLOPs it removes. ToMe also matches within each window but on the keys averaged over the $H$ heads, which adds $\mathcal{O}(WDN_l/H)$ at every block. ALGM adds $\mathcal{O}(N_0D)$ for its local merging, which compares only the tokens within each small window, and $\mathcal{O}(N_3^2D)$ for its global merging, which compares all tokens of the third stage with each other.

\subsection{Latency of the Overhead Operations}
\label{app:breakdown_ops}

Table~\ref{tab:breakdown} splits the overhead latency of each method into scoring (Appendix~\ref{app:breakdown_flops}) and three operations that only move the token features: gather/reduce, scatter, and sort. The overhead thus consists of both computation and memory traffic, and its latency depends on how often each operation runs, not only on its FLOPs.

\textbf{Data movement.} \textit{Gather/reduce} builds the reduced token set. The merging methods accumulate each group of source tokens into its destination with atomic additions, as in their official implementations~\citep{tome, algm, gitmerge3d}. Many source tokens add into the same destination at once, and the GPU serializes these conflicting writes. SpaceCache instead gathers the cache-miss tokens into a contiguous subset, reading each token once without any accumulation. \textit{Scatter} restores the full token layout. The merging methods copy each merged feature back to its source tokens. SpaceCache writes the newly computed features of the cache-miss tokens and the cached features of the cache-hit tokens to their own positions. \textit{Sort} maintains the cache and is used only by SpaceCache. The Commit Gate merges the new entries into the table, which is kept sorted by region key for the binary search, and evicts the expired and low-confidence entries, once per stage and frame.

\textbf{Repetition across blocks.} SpaceCache runs each operation once per stage. In contrast, GitMerge3D repeats its scoring, gather/reduce, and scatter at every block. ToMe repeats its scoring and gather/reduce at every block but scatters only at the decoder unpooling. ALGM runs each operation only twice, for its local merging at stage 0 and its global merging at stage 3.

\subsection{Memory Overhead}
\label{app:breakdown_memory}

Table~\ref{tab:breakdown} splits the peak memory into the \textit{model} weights (602\,MiB for every method), the \textit{activation} memory (the peak of the intermediate tensors of the forward pass), and the \textit{cache} (the persistent table of SpaceCache). SpaceCache$_{\text{Stage-}3}$ and SpaceCache$_{\text{Stage-}3,4}$ add 18 and 26\,MiB of activations, because they hold the newly computed features of the cache-miss tokens until the Commit Gate after the decoder. Each cache entry holds the region key (64-bit integer), the feature (single precision, 432 channels at the third stage and 576 at the fourth), and the age of one region. The cache takes 47.7\,MiB for the third stage and 21.7\,MiB for the fourth (69.4\,MiB in total). GitMerge3D restores the tokens right after each attention and keeps nothing across blocks, so its activation memory stays within 2\,MiB of the backbone. ToMe and ALGM instead keep their merge maps and the unmerged tokens until the decoder restores them, which raises the activation memory by 175 and 58\,MiB.

\begin{table}[b!]
\centering
\footnotesize
\setlength{\tabcolsep}{4pt}
\caption{\textbf{FLOPs, latency, and memory breakdown} on the representative scenes, each method at its point within a 2.0-point mIoU drop. The latency and memory blocks are plotted in Figure~\ref{fig:breakdown}. FLOPs and latency are per frame and use dense attention, and memory uses flash-attention.}
\label{tab:breakdown}
\begin{tabular}{@{}lrrrrrr@{}}
\toprule
& \multicolumn{6}{c}{Waymo with Utonia} \\
\cmidrule(l){2-7}
& \textbf{Utonia} & + ToMe$^{*}$ & + ALGM$^{*}$ & + GitMerge3D$^{\dagger}$ & \makecell[r]{+ SpaceCache\\Stage-3} & \makecell[r]{+ SpaceCache\\Stage-3,4} \\
Sweep parameter & -- & $r{=}0.025$ & $\tau{=}0.98$ & $r{=}0.60$ & $\theta{=}0.55$ & $\theta{=}0.55$ \\
\midrule
\multicolumn{7}{@{}l}{\textbf{Accuracy}} \\
\quad mIoU (\%)$\uparrow$ & 67.17 & 65.36 & 65.25 & 65.21 & 65.20 & 65.44 \\
\midrule
\multicolumn{7}{@{}l}{\textbf{GFLOPs}} \\
\quad Linear and SpConv & 3321.5 & 2485.2 & 2622.8 & 3321.5 & 2178.0 & 1955.5 \\
\quad Attention products & 1116.2 & 928.5 & 951.8 & 178.9 & 889.7 & 857.3 \\
\quad Saving & -- & $-$1024.0 & $-$863.1 & $-$937.3 & $-$1370.0 & $-$1625.0 \\
\quad Overhead & -- & 13.4 & 87.2 & 124.0 & ${<}$0.001 & ${<}$0.001 \\
\cmidrule(l){1-7}
\quad Total$\downarrow$ & 4437.7 & 3427.1 & 3661.8 & 3624.4 & 3067.7 & 2812.7 \\
\quad $\Delta$FLOPs (\%)$\downarrow$ & -- & $-$22.8 & $-$17.5 & $-$18.3 & $-$30.9 & $-$36.6 \\
\midrule
\multicolumn{7}{@{}l}{\textbf{Latency (ms)}} \\
\quad SpConv & 188.1 & 156.9 & 161.6 & 188.0 & 131.0 & 120.2 \\
\quad Attention & 340.6 & 301.9 & 304.2 & 184.4 & 272.0 & 261.9 \\
\quad FFN + others & 171.9 & 149.7 & 163.6 & 173.3 & 144.2 & 139.6 \\
\quad Overhead & -- & 40.0 & 11.6 & 136.4 & 2.6 & 4.8 \\
\qquad Scoring & -- & 9.1 & 8.6 & 41.1 & 0.3 & 0.5 \\
\qquad Gather/reduce & -- & 30.5 & 2.8 & 89.8 & 0.4 & 0.8 \\
\qquad Scatter & -- & 0.4 & 0.2 & 5.5 & 0.1 & 0.2 \\
\qquad Sort & -- & -- & -- & -- & 1.9 & 3.4 \\
\cmidrule(l){1-7}
\quad Latency$\downarrow$ & 700.6 & 648.4 & 641.0 & 682.0 & 549.9 & 526.5 \\
\quad Speedup ($\times$)$\uparrow$ & 1.00 & 1.08 & 1.09 & 1.03 & 1.27 & 1.33 \\
\midrule
\multicolumn{7}{@{}l}{\textbf{Peak memory (MiB)}} \\
\quad Weights & 602 & 602 & 602 & 602 & 602 & 602 \\
\quad Activation & 758 & 934 & 817 & 756 & 776 & 784 \\
\quad Cache & -- & -- & -- & -- & 48 & 69 \\
\cmidrule(l){1-7}
\quad Peak$\downarrow$ & 1360 & 1535 & 1419 & 1358 & 1426 & 1455 \\
\quad $\Delta$Peak (\%)$\downarrow$ & -- & 12.9 & 4.3 & $-$0.1 & 4.8 & 7.0 \\
\bottomrule
\end{tabular}
\end{table}

\end{document}